\documentclass{article}

\PassOptionsToPackage{numbers, compress}{natbib}
\usepackage[preprint]{neurips_2026}

\usepackage[utf8]{inputenc} 
\usepackage[T1]{fontenc}    
\usepackage{hyperref}       
\usepackage{url}            
\usepackage{booktabs}       
\usepackage{amsfonts}       
\usepackage{nicefrac}       
\usepackage{microtype}      
\usepackage{xcolor}         

\usepackage[]{graphicx}       
\graphicspath{{img/}}     
\usepackage{amsmath}
\usepackage{framed}

\newboolean{draftenabled}
\setboolean{draftenabled}{true} 

\ifthenelse{\boolean{draftenabled}}
{
    \newenvironment{draft}{
       
       \MakeFramed{\advance\hsize-\width\FrameRestore}}
     {\endMakeFramed}

    \newcommand{\draftinline}[1]{\colorbox{pink}{#1}}
}{
    \newcommand{\draftinline}[1]{}

}

\definecolor{SkyBlue}{rgb}{0.66,0.88,0.93}

\newcommand{\bx}[0]{\mathbf x}
\newcommand{\ie}[0]{\textit{i.e.}}
\newcommand{\eg}[0]{\textit{e.g.}}

\title{Implicit Data Synthesis for Contrastive Unsupervised Data Augmentation}

\author{%
    Patrick Kage \\
    School of Informatics\\
    The University of Edinburgh\\
    Edinburgh, Scotland EH8 9AB UK\\
    \texttt{p.kage@ed.ac.uk} \\
    \And
    Trevor Hedges \\
    Massachusetts Institute of Technology Lincoln Laboratory \\
    Lexington, MA 02421 USA\\
    \texttt{trevor.hedges@ll.mit.edu} \\
    \And
    N. Siddharth \\
    School of Informatics\\
    The University of Edinburgh\\
    Edinburgh, Scotland EH8 9AB UK\\
    \texttt{n.siddharth@ed.ac.uk} \\
    \And
    Pavlos Andreadis \\
    School of Informatics\\
    The University of Edinburgh\\
    Edinburgh, Scotland EH8 9AB UK\\
    \texttt{pavlos.andreadis@ed.ac.uk}
}

\begin{document}

\maketitle

\begin{abstract}

    Scientific observations generate large quantities of unlabeled data which
    is laborious to hand-label, making unsupervised learning techniques
    valuable for processing datasets. Among these approaches, contrastive
    learning provides a convenient mechanism for extracting structural
    representations from unannotated datasets. For natural imagery, the general
    approach is to use a variety of data-space augmentation methods in order to
    generate synthetic samples; however, for scientific observations data-space
    perturbations can fundamentally alter the underlying data. Our proposed
    method is to generate contrastive samples by perturbing the network weights
    rather than the underlying data, thus more closely preserving the structure of
    the data. We demonstrate this technique using a SimCLR-based pipeline
    applied over radar observations of meteors, and
    show performance gains under matched protocols.

\end{abstract}


\section{Introduction}


The advent of modern machine learning techniques provides scientists powerful
tools for the processing of experiment data, allowing for analyses such as
classification, clustering, anomaly detection, semantic segmentation, and more.
However, the application of such tools is limited by the core drawback of
supervised learning: the necessity of large quantities of labeled data with
which to train machine learning models. In scientific domains, the creation of
large, high-quality datasets is extremely labor-intensive, requiring domain
experts to tediously label thousands of samples by hand. To combat this, the
field of \textit{unsupervised learning} has emerged, which allows for label-free
analysis of datasets---skirting the need for human labeling by leveraging the
inherent properties of the data to create an analysis pipeline. Within this
field, \textit{unsupervised representation learning} is a common task, which
aims to distill high-dimensional data into low-dimensional vectors capturing
semantic information about the sample, allowing for easier downstream
analysis~\cite{pl-review, SimCLR}.


Contrastive learning techniques are a natural fit for creating unsupervised
representation learning systems as they strive to learn common
features between similar samples while separating dissimilar samples. This yields
a system which is able to cluster semantically similar samples in feature space
without needing a class assignment---only a metric of how similar samples are
to each other~\cite{contrastive-survey}. In general, these systems work by
augmenting a sample \(\bx\) to create a set of (typically two) positive samples\footnote{Though
pairwise contrastive losses are common, others are available---see
\eg~\cite{scale-alibi} and~\cite{SwAV}.} \(\{\hat\bx_1,
\hat\bx_2, \dots\}\) with the property that these two samples share the same
underlying class assignment. These samples are then projected into latent space
by an encoder network and projection network which attempts minimize the space
between positive samples and the rest of the batch (negative
samples)~\cite{contrastive-survey, SimCLR}, yielding a system capable of
semantically embedding unlabeled samples into a latent space.


A core component of the success of contrastive learning techniques is the data
augmentation strategy, \ie~the ability for the system to generate positive
samples \(\hat\bx\)  that differ from the original sample \(\bx\), but
map to the same location in the final embedding space. It has been
shown that effective augmentations are key to a robust embedding
system~\cite{contrastive-survey}. This brings us to the key problem addressed
by this paper: while many strategies exist for natural
imagery~\cite{augmentation-survey}, creating augmentations for non-image data
requires bespoke augmentations which take into account what features of the
data can be changed without fundamentally altering its class---a challenging
task.


This difficulty is particularly pronounced when analyzing meteor data from 
scientific radars such as the Jicamarca Radio Observatory (JRO) in Lima, Peru.
These extremely sensitive radar facilities can detect thousands of tiny meteors
(originating from dust particles in space) entering Earth's atmosphere per hour~\cite{hedges_meteor_2022},
yielding enormous datasets.
The most common meteor radar signatures observed by JRO, known as \textit{head echoes},
provide insights into atmospheric composition and formation of the solar system, yet the
process of manually labeling radar data is prohibitively
time-consuming~\cite{hedges_meteor_2024}. Furthermore, standard natural imagery
augmentations cannot be trivially applied to radar data without corrupting
the underlying physical phenomena represented in the signal.
Radar observations of meteors are processed as two-dimensional datasets where the horizontal axis represents time and the vertical axis altitude, so the spatial axes are not interchangeable. A meteor's apparent velocity is given by the slope of its trajectory in this plane: arbitrary rotations rescale that velocity, and reflections invert its sign, producing samples in which meteors appear to ascend through the atmosphere. Pixel intensities are similarly tied to calibrated signal-to-noise ratios that represent the size of a meteor rather than perceptual quantities, so color jitter augmentations are not physically meaningful either.


To address this, we propose using \textit{implicit data synthesis} (IDS).
Rather than modifying the data in data space, we instead add a bounded noise
term to a subset of the layers of the encoder network, effectively augmenting
the data in representation space. This approach has previously shown efficacy
in anomaly detection via Wasserstein-Distributed Outlier Exposure
(W-DOE)~\cite{wdoe}, where a similar approach to IDS was utilized to expand the
out-of-distribution dataset. In this work, we adapt this theory to contrastive
learning, creating a robust unsupervised representation learning pipeline
applicable to meteor radar data and other scientific observations. We show that
IDS alone is as effective as data-space augmentations for downstream tasks in
our dataset, and show that IDS is also effective in natural imagery domains by
demonstrating performance over CIFAR-10~\cite{cifar10}.
 

In summary, our paper contributions are:

\begin{enumerate}
    \item We adapt implicit data synthesis to contrastive learning,
        replacing data-space augmentation with per-forward-pass
        perturbation of a single fully-connected layer in the encoder.

    \item We evaluate the method on synthetic meteor radar
        observations, a setting in which canonical SimCLR augmentations
        corrupt the underlying physical signal, and show that IDS matches
        or outperforms a flip-and-rotation baseline across CNN encoders.

    \item We verify that the technique generalizes beyond the radar domain by
        reproducing the comparison on CIFAR-10 under a matched
        (flip-and-rotation) augmentation budget, and characterize the
        sensitivity of the method to the perturbation scale~\(s_l\).

\end{enumerate}

\section{Related work}\label{sec:related}

Modern
self-supervised learning for visual representations is dominated by
joint-embedding methods that learn invariances by aligning two views of the
same sample. SimCLR~\cite{SimCLR} establishes the canonical recipe of random
crop, color jitter, and Gaussian blur trained against an NT-Xent loss;
MoCo~\cite{moco} replaces the in-batch negatives with a momentum-encoded queue;
BYOL~\cite{byol} and SimSiam~\cite{simsiam} dispense with negatives entirely,
relying on a predictor network and asymmetric stop-gradient to avoid
representational collapse. More recent work shifts to self-distillation with
vision transformers (\eg~the DINO family~\cite{dino,dinov2,dinov3}) and to masked
reconstruction (\eg~MAE~\cite{mae}). Across all of these methods the
view-generation step is essentially unchanged from SimCLR: positive pairs are
produced by stochastic data-space augmentation. Our contribution is orthogonal
to the choice of objective---the perturbation we introduce in
Section~\ref{sec:methodology} can in principle be plugged into any of these
frameworks---and we adopt the SimCLR setting because it isolates the
view-generation mechanism most cleanly.

Mechanically, IDS resembles a long line
of stochastic-forward-pass methods. Dropout~\cite{dropout} and DropConnect
\cite{dropconnect} inject multiplicative Bernoulli noise into activations and
weights respectively; stochastic depth \cite{stochasticdepth} randomly skips
entire residual blocks; and Bayesian treatments such as MC
dropout~\cite{mcdropout} and SWAG \cite{swag} interpret stochastic forward
passes as samples from an approximate weight posterior, used at inference time
for uncertainty quantification. In all of these, the noise serves either as a
regularizer during training or as a posterior sampling mechanism at test time;
the loss is computed on a single noisy forward pass.

IDS uses the same mechanism for a different purpose: two \emph{independent}
stochastic forward passes on the same input produce a positive pair for a
contrastive loss, and the noise is the sole source of view variation rather
than a regularizer layered on top of data-space augmentation. The closest
precedent is SimCSE \citep{simcse}, which uses independently sampled dropout
masks to construct positive pairs for contrastive sentence
embedding---demonstrating that stochastic forward passes \emph{can} substitute
for data-space augmentation in a contrastive setting, but doing so via
Bernoulli activation masks on a Transformer backbone, in a domain (text) where
the analogue of crop-and-flip is poorly defined to begin with. We extend this
idea to image contrastive learning with additive Gaussian perturbation of
weights at a designated layer, motivated by scientific imaging domains in which
the standard data-space augmentations corrupt the underlying physical signal.

A natural comparison is also the variational autoencoder~\cite{vae}, whose encoder
defines a distribution $q(z \mid x)$ from which multiple latents can be sampled
for a single input. IDS differs in where the stochasticity lives---we sample a
random encoder $f_{\theta+\lambda}$ from a fixed distribution over functions
rather than sampling latents from a learned, input-dependent distribution---and
has no decoder, reconstruction term, or KL penalty shaping the perturbation
scale.

\section{Methodology}\label{sec:methodology}
 
Implicit data synthesis is implemented by adding a noise term to the standard
formulation for a Multilayer Perceptron (MLP):
 
\begin{equation}
    \bold h^{(l)}=\sigma\big((W^{(l)} + \bold\lambda^{(l)})\bold h^{(l-1)}\big)
    \label{eq:ids}
\end{equation}
 
where \(\bold h^{(l)}\) is layer \(l\)'s activation, \(\bold h^{(l-1)}\) is the
previous layer's activation, \(W^{(l)}\) are layer \(l\)'s weights, and \(\bold
\lambda^{(l)}\) is a perturbation matrix of the same shape as \(W^{(l)}\). The
entries of \(\bold\lambda^{(l)}\) are sampled i.i.d.~from \(\mathcal N(0,
s_l)\), where \(s_l\) is a per-layer scale hyperparameter. The perturbation is
resampled on every forward pass; the underlying weights \(W^{(l)}\) are
unaffected and are updated only through the contrastive loss. The geometric
intuition for this is that perturbations in weight space induce a distribution
over functions, which induces a distribution over the representations of the
samples.

The weight-perturbation operator in Equation~\ref{eq:ids} is adapted from
Wasserstein-Distributed Outlier Exposure (W-DOE)~\cite{wdoe}, which uses
perturbations of network weights to synthesize auxiliary
out-of-distribution (OOD) samples for outlier exposure training. In W-DOE, the
perturbation expands the support of an auxiliary OOD set used in a
supervised binary in-distribution-vs-OOD objective; the perturbed forward
passes serve as additional negative examples whose distance from the
in-distribution manifold is bounded by the Wasserstein radius induced by
\(s_l\). We retain the perturbation mechanism but apply it in a fundamentally
different role: rather than generating negatives against a fixed
in-distribution density, two independent draws of \(\bold\lambda\) are used to generate a
positive pair for a self-supervised contrastive objective. The two settings
share the geometric intuition that small weight perturbations induce
bounded movements in representation space, but they differ in what that
bounded movement is used for---bounding the OOD synthesis radius in W-DOE,
and bounding the intra-class spread of synthesized positives.
Consequently, the W-DOE convergence and bound results, which are stated for
the OOD-exposure objective, do not transfer directly to the NT-Xent
setting, and we treat the perturbation as a heuristic augmentation operator
rather than one with inherited theoretical guarantees.

\begin{figure}
    \centering
    \includegraphics[width=0.9\textwidth]{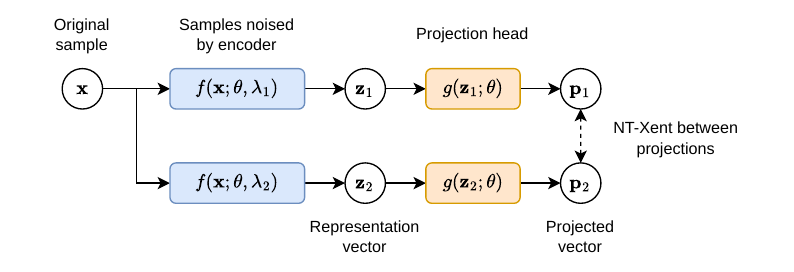}
    \caption{Diagram showing the contrastive IDS architecture. Note that, as opposed to SimCLR, there is no data-space augmentation block---rather, the augmentations are derived from the noise term \(\lambda\) parameterizing the encoder network \(f(\,\cdot\,)\).}
    \label{fig:simclr-change}
\end{figure}

Within the SimCLR framework, this perturbation provides the mechanism for
constructing positive pairs. Given an input \(\bx\), two independent draws
\(\bold\lambda^{(l)}_1, \bold\lambda^{(l)}_2 \sim \mathcal N(0, s_l)\) yield
two perturbed encoders \(f_{\bold\lambda_1}\) and \(f_{\bold\lambda_2}\), and
the positive pair is \((\hat\bx_1, \hat\bx_2) = (f_{\bold\lambda_1}(\bx),
f_{\bold\lambda_2}(\bx))\). The two views are then projected and scored under
the standard NT-Xent loss in the same manner as SimCLR; only the source of
view variation has changed. Because the input \(\bx\) is passed through the
network unmodified, no data-space augmentation is required, and the
augmentation does not need to be designed against the physical structure of
the data. The resulting configuration is shown in
Figure~\ref{fig:simclr-change}.
 
The \(\mathbf\lambda^{(l)}\) perturbation is not applied uniformly across the network. In practice,
\(s_l\) is set to zero for all but a small number of layers, restricting the
weight-space augmentation to a single, well-defined site. The choice of which
layers to perturb is left as a design parameter. In this work, the perturbation
is applied at the first fully-connected layer following the convolutional
stack (see Section~\ref{sec:results}). Restricting perturbation to a
representation-space layer rather than to the convolutional features keeps the
augmentation acting on already-abstracted features.  Perturbing in feature space yields
samples that remain close to the data manifold, whereas perturbing earlier
layers risks destroying low-level structure that the encoder has not yet
learned to ignore.

The scale parameter \(s_l\) controls the effective spread of the synthesized
positives. With \(s_l = 0\) the two views collapse to the same point and the
contrastive loss becomes degenerate: as \(s_l\) grows the two views become
arbitrarily dissimilar and the encoder is asked to embed unrelated points
together, which destabilizes training. \(s_l\) thus plays a role analogous to
augmentation strength in standard SimCLR, and we treat it as a hyperparameter
to be swept (Section~\ref{sec:results}). At evaluation time, we set \(s_l =
0\) in all layers, recovering a deterministic encoder.

\section{Experimental Setup}

\begin{figure}
    \centering
    \includegraphics[width=0.99\textwidth]{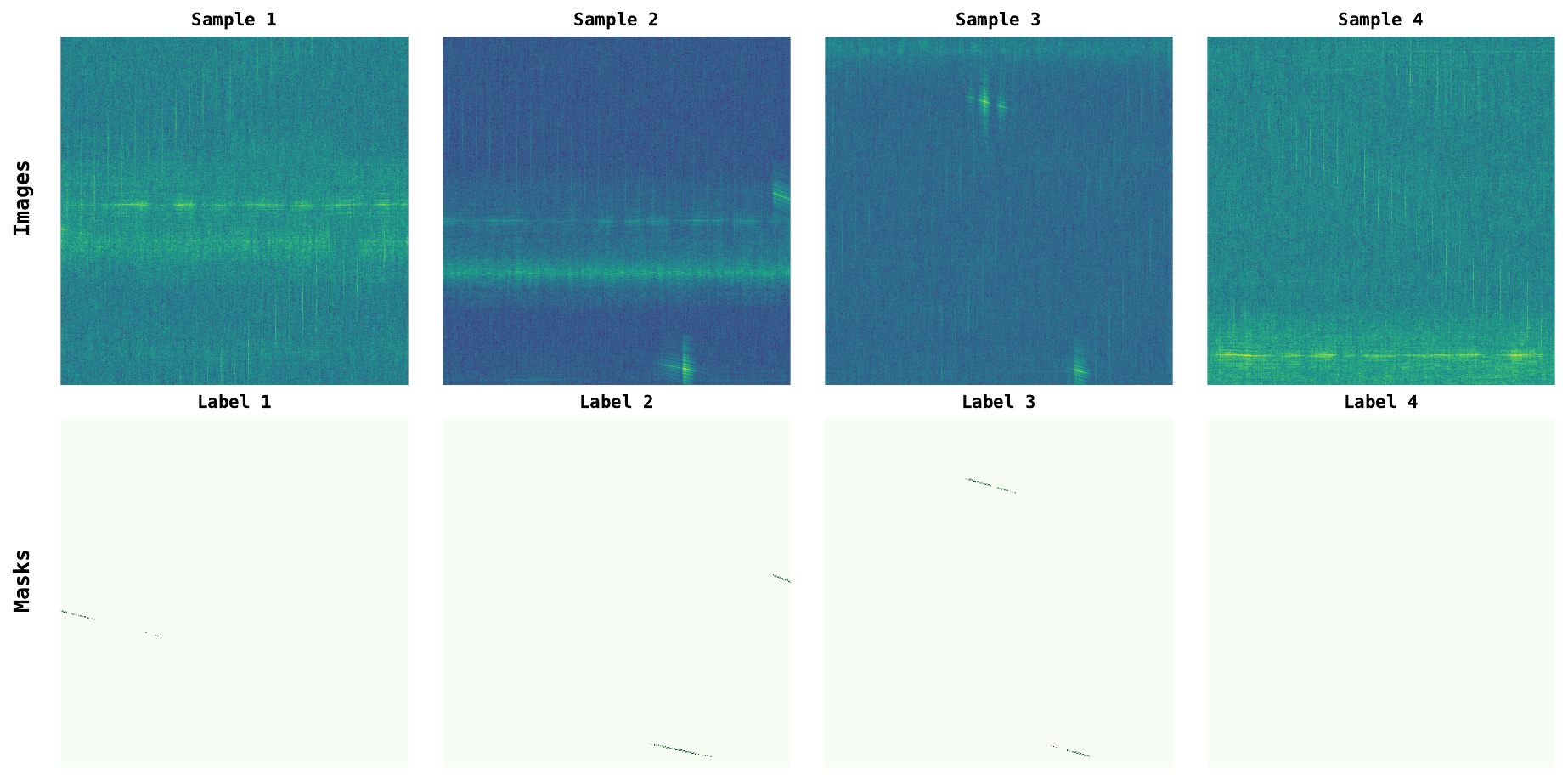}
    \caption{A selection of synthetic radar data samples and associated per-pixel labels indicating meteor locations.}
    \label{fig:synthetic_sample}
\end{figure}

To validate the efficacy of implicit data synthesis within the contrastive
framework, we utilize a dataset of synthetically-generated radar samples
containing simulated meteor head echo observations. Since radar observations of
meteor head echoes can be simulated in a computationally efficient manner via various
physical approximations that preserve realism~\cite{hedges_meteor_2024},
this synthetic data closely mirrors real-world distributions while providing
a high-volume testbed for our representation learning pipeline. Examples of
synthetic radar data samples, along with the pixel-wise labels indicating meteor head echoes,
are shown in Figure~\ref{fig:synthetic_sample}.

We crop down the synthetic observations (originally \(512\times 512\) images) to a set
of \(N\times N\) subtiles, allowing us to evaluate the spatial resolution
bounds of the learned representations and assess how varying amounts of spatial
context affect the contrastive learning process.

To establish a performance baseline, models are trained utilizing standard
data-space augmentations. Specifically, random horizontal/vertical flips and
rotations. This baseline represents the standard approach in the literature and
serves as the comparative benchmark against our proposed implicit data
synthesis technique.

\subsection{Evaluation}\label{sec:evaluation}

To evaluate the quality of the learned representations, we freeze the encoder
weights post-training, set the \(s_l\) to zero in all layers, and remove the
projection head. We train both a shallow MLP classifier (a linear probe)
as well as a \(k\)-nearest-neighbor classifier on the resulting embeddings,
evaluated against a binary classification problem: determining whether a meteor
head echo exists in a given sample. The performance of this probe serves as our
primary metric for the semantic richness of the unsupervised latent space.
Comparing the \(k\)-NN and MLP performance allows us to evaluate both the local
clustering and the broader linear separability of the learned representations.

\subsection{Model Architectures}\label{sec:model-architectures}

We evaluate the representation learning pipeline across a spectrum of encoder
architectures to assess both scalability and capacity requirements. The
evaluated base models include a series of simple Convolutional Neural Networks
(CNNs) explicitly scaled for varying input tile sizes ($8\times 8$ through
$256\times 256$) paired with MLP layers (which receive the noise), as well as
standard ResNet-18 and ResNet-50 architectures. The ResNets are evaluated
exclusively on the $256\times 256$ tiles, and again only the fully-connected MLP layers receive noise.

All CNN encoders follow a common template: a stack of
\texttt{conv2d}--ReLU--\texttt{maxpool2d} blocks followed by a two-layer MLP
that produces a representation vector of dimension $D_r = 32$. The number of
convolutional blocks, the kernel sizes, and the pooling strides are scaled with
the input tile size so that the spatial extent collapses to a $3\times 3$
feature map prior to flattening. The smallest two variants (CNN-$8\times 8$ and
CNN-$16\times 16$) use two convolutional blocks with $2\times 2$ pooling; the
mid-sized variants (CNN-$32\times 32$ through CNN-$64\times 64$) retain two
blocks but increase the kernel size and pooling stride; and the largest
variants (CNN-$128\times 128$ and CNN-$256\times 256$) use three convolutional
blocks. Channel widths grow with depth, ranging from $16$--$32$ channels in the
smallest model up to $32$--$64$--$128$ in the largest. Across all CNN variants
the channel counts and the MLP hidden sizes are kept small relative to the
ResNet baselines, in order to isolate the effect of input resolution from raw
model capacity.

For all encoders---CNN and ResNet alike---the standard SimCLR projection head
is attached to the representation vector: a two-layer MLP with a ReLU
nonlinearity that maps from $\mathbb R^{D_r}$ to a projection space of dimension
$D_p = 16$. For the ResNet backbones, the default classification head is
replaced with an identity layer so that the backbone outputs its native feature
dimension ($512$ for ResNet-18, $2048$ for ResNet-50) directly into the
projection MLP. The contrastive loss is computed in projection space, while
the downstream linear probe and $k$-NN classifier operate on the
representation vector $\mathbf h \in \mathbb R^{D_r}$ obtained immediately
before the projection head.

When implicit data synthesis is enabled, the perturbation $\boldsymbol\lambda$
is applied only to the weights of the first fully-connected layer of the
encoder's MLP head (\ie, \texttt{fc1} in the CNN variants). All convolutional
weights and the projection head remain unperturbed. This restricts the
weight-space augmentation to a single, well-defined site in the network and
keeps the perturbation budget comparable across architectures of differing
depth.

\subsection{Baselines}\label{sec:baselines}

To assess the efficacy of implicit data synthesis as a contrastive
augmentation strategy, we compare against three baseline configurations,
designed to isolate both the in-domain performance on meteor radar data and the
generality of the technique across data modalities: a direct application of
natural imagery augmentation to the radar dataset, an in-domain augmentation
applied to the radar dataset, and a comparison to CIFAR-10 with both a matching set of 
baseline augmentations and IDS applied.

The first two baselines target our primary dataset directly. First, we train a
SimCLR pipeline on the synthetic meteor observations using a restricted
set of data-space augmentations: random horizontal/vertical flips and
rotations. We deliberately omit the remainder of the canonical SimCLR
augmentation suite~\cite{SimCLR}---in particular, color jitter and aggressive
cropping---as these operations do not compare cleanly against implicit data
synthesis on radar data because they would corrupt the underlying physical
signal rather than produce a semantically equivalent view. This configuration
provides a direct point of comparison against our proposed method, in which the
same encoder is trained using only implicit data synthesis (\ie, weight-space
perturbation) as the source of positive-pair variation. The contrast between
these two configurations isolates the effect of replacing data-space
augmentations with representation-space perturbations on a domain in which
natural imagery augmentations are not physically well-motivated.

Additionally, we evaluate an in-domain, physics-motivated
augmentation strategy tailored to the structure of meteor radar observations. Each
synthetic sample is transformed via a column-wise FFT, and a contiguous band of
the resulting spectrum (set to \(15\%\)) is zeroed out before the inverse transform is applied to
recover a perturbed sample in the original domain. The masked band is selected
randomly per sample, yielding two distinct views when applied twice to the same
input. This augmentation is used as the sole source of positive-pair variation;
no flips, rotations, or other data-space perturbations are composed with it.
The motivation is to produce views that differ in their frequency content while
preserving the spatial and temporal structure of the underlying radar signal,
providing a domain-appropriate point of comparison against both the
flip-and-rotation baseline and implicit data synthesis.

The remaining two baselines are designed to demonstrate that implicit data
synthesis generalizes beyond radar observations to natural imagery. To this end,
we repeat the above pair of experiments on the CIFAR-10~\cite{cifar10} dataset:
one configuration trained using the same flip-and-rotation augmentation set,
and a second trained using implicit data synthesis in place of those
augmentations. To maintain parity with the binary classification protocol used
for the meteor radar evaluation, we collapse the ten CIFAR-10 classes into two
superclasses: \textit{vehicle} (airplane, automobile, ship, truck) and
\textit{animal} (bird, cat, deer, dog, frog, horse). CIFAR-10 thus serves as a
natural imagery testbed where the contrastive learning protocol is
well-characterized, and allows us to verify that the gains observed on meteor
data are not an artifact of the radar domain.

Across all of the above configurations, we hold the encoder architecture,
projection head, optimizer, batch size, and number of training epochs fixed,
varying only the augmentation strategy. Downstream evaluation follows the
protocol described in Section~\ref{sec:model-architectures}: encoder weights are frozen,
\(s_l\) is set to zero, the projection head is removed, and both a linear probe
and a $k$-NN classifier are trained on the resulting embeddings.

\section{Results}\label{sec:results}

\begin{figure}
    \centering
    \includegraphics[width=0.99\textwidth]{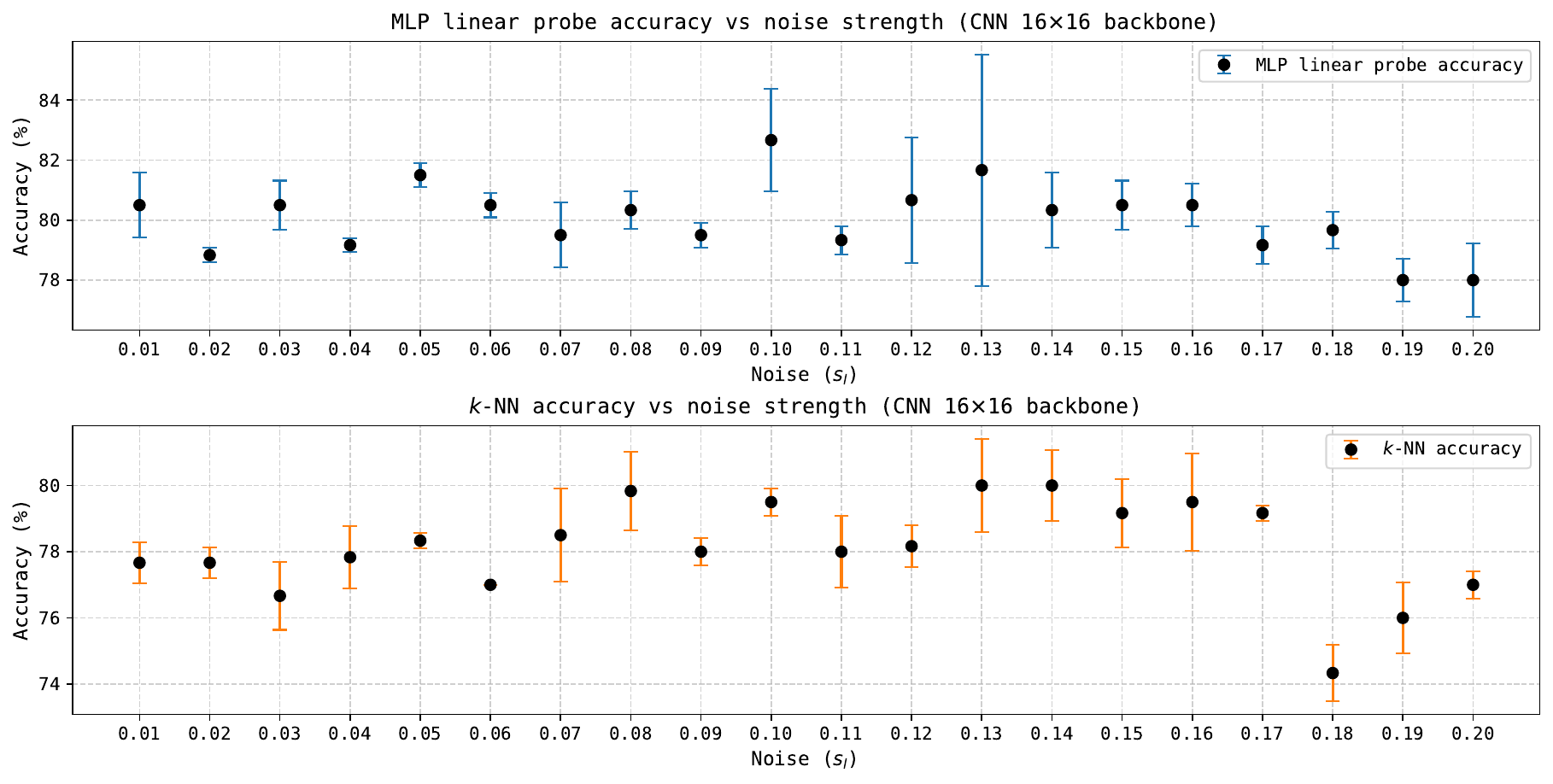}
    \caption{Noise sweep across CNN-16\(\times\)16 architecture reporting accuracy on meteor classification with both the MLP linear probe and \(k\)-NN accuracy, evaluated as in Section~\ref{sec:model-architectures}. All models were run 5 times for an identical number of epochs. Mean and standard deviation are reported.}\label{fig:sweep-cnn16x16}
\end{figure}

We report linear-probe and \(k\)-NN classification accuracy across three
augmentation conditions: the flip-and-rotation baseline, the in-domain FFT
masking augmentation, and IDS. Each condition is evaluated on the synthetic
meteor dataset across the six CNN variants and the two ResNet backbones
described in Section~\ref{sec:model-architectures}, and on CIFAR-10 across the
same encoder set. All runs are repeated at minimum three times with different
seeds (varying per experiment); we report the mean and standard deviation of
the resulting accuracies. The sensitivity of IDS to \(s_l\) is examined
separately on CNN-16\(\times\)16 in Figure~\ref{fig:sweep-cnn16x16}.

\begin{table}[htbp]
    \caption{Top-line model accuracy on meteor classification across baseline, in-domain (FFT), and IDS noising schemes, with the best performance bolded. The classification task was evaluated as described in Section~\ref{sec:evaluation}. All models were run 3 times with \(s_l = 0.02\) for an identical number of epochs. Mean and standard deviation are reported.}
    \label{tab:topline}
    \centering
    \begin{tabular}{l l l l}
        \toprule
        \multicolumn{4}{c}{MLP (\%)} \\
        \midrule
        Model & \multicolumn{1}{c}{Baseline} & \multicolumn{1}{c}{In-domain} & \multicolumn{1}{c}{IDS} \\


        \midrule

        CNN-8\(\times\)8 & 
            \(83.50 \pm 1.41\) &
            \(80.67 \pm 1.18\) &
            \(\mathbf{84.50 \pm 0.41}\) \\

        CNN-16\(\times\)16 &
            \(80.83 \pm 1.65\) &
            \(71.67 \pm 1.31\) &
            \(\mathbf{85.17 \pm 1.43}\) \\

        CNN-32\(\times\)32 & 
            \(77.33 \pm 0.85\) &
            \(74.67 \pm 1.65\) &
            \(\mathbf{78.00 \pm 0.00}\) \\

        CNN-64\(\times\)64 &
            \(75.83 \pm 0.24\) &
            \(74.33 \pm 1.03\) &
            \(\mathbf{76.67 \pm 0.47}\) \\

        CNN-128\(\times\)128 &
            \(\mathbf{75.17 \pm 0.47}\) &
            \(51.67 \pm 0.94\) &
            \(73.83 \pm 1.25\) \\

        CNN-256\(\times\)256 &
            \(66.50 \pm 1.22\) &
            \(52.83 \pm 0.62\) &
            \(\mathbf{71.17 \pm 0.24}\) \\

        ResNet-18 &
            \(87.17 \pm 0.62\) &
            \(\mathbf{93.67 \pm 1.03}\) &
            \(86.50 \pm 0.41\) \\

        ResNet-50 &
            \(92.17 \pm 0.24\) &
            \(93.67 \pm 0.62\) &
            \(\mathbf{94.33 \pm 0.47}\) \\

        \midrule
        \multicolumn{4}{c}{ \(k\)-NN (\%)} \\
        \midrule
        Model & \multicolumn{1}{c}{Baseline} & \multicolumn{1}{c}{In-domain} & \multicolumn{1}{c}{IDS} \\
        \midrule

        CNN-8\(\times\)8 & 
            \(\mathbf{84.50 \pm 0.41}\) &
            \(80.67 \pm 0.62\) &
            \(84.00 \pm 0.41\) \\

        CNN-16\(\times\)16 &
            \(77.83 \pm 0.47\) &
            \(75.83 \pm 1.65\) &
            \(\mathbf{81.67 \pm 0.24}\) \\

        CNN-32\(\times\)32 & 
            \(76.67 \pm 2.01\) &
            \(73.83 \pm 0.24\) &
            \(\mathbf{77.17 \pm 0.94}\) \\

        CNN-64\(\times\)64 &
            \(77.33 \pm 0.85\) &
            \(79.00 \pm 2.12\) &
            \(\mathbf{79.67 \pm 2.01}\) \\

        CNN-128\(\times\)128 &
            \(75.83 \pm 1.03\) &
            \(54.17 \pm 0.62\) &
            \(\mathbf{83.33 \pm 1.03}\) \\

        CNN-256\(\times\)256 &
            \(72.33 \pm 0.85\) &
            \(57.67 \pm 6.60\) &
            \(\mathbf{80.17 \pm 0.94}\) \\

        ResNet-18 &
            \(82.50 \pm 2.12\) &
            \(\mathbf{90.17 \pm 0.47}\) &
            \(80.33 \pm 2.25\) \\

        ResNet-50 &
            \(88.33 \pm 0.62\) &
            \(\mathbf{92.50 \pm 0.82}\) &
            \(86.00 \pm 1.08\) \\

        \bottomrule
    \end{tabular}
\end{table}

\begin{table}[htbp]
    \caption{Accuracy over CIFAR-10 across baseline and IDS noising schemes, with the best performance bolded. See Section~\ref{sec:baselines} for experimental configuration. All models were run 3 times with \(s_l = 0.02\) for an identical number of epochs.}
    \label{tab:cifar10}
    \centering
    \begin{tabular}{l l l}
        \toprule
        \multicolumn{3}{c}{MLP (\%)} \\
        \midrule
        Model & \multicolumn{1}{c}{Baseline} & \multicolumn{1}{c}{IDS} \\

        \midrule

        CNN-8\(\times\)8 & 
            \(66.50 \pm 0.82\) &
            \(\mathbf{74.33 \pm 1.55}\) \\

        CNN-16\(\times\)16 &
            \(62.17 \pm 0.62\) &
            \(\mathbf{71.17 \pm 0.85}\) \\

        CNN-32\(\times\)32 & 
            \(62.83 \pm 1.65\) &
            \(\mathbf{74.33 \pm 3.40}\) \\

        CNN-64\(\times\)64 &
            \(62.83 \pm 0.62\) &
            \(\mathbf{71.17 \pm 0.62}\) \\

        CNN-128\(\times\)128 &
            \(61.67 \pm 0.62\) &
            \(\mathbf{68.67 \pm 0.24}\) \\

        CNN-256\(\times\)256 &
            \(64.67 \pm 2.36\) &
            \(\mathbf{70.33 \pm 1.18}\) \\

        ResNet-18 &
            \(86.17 \pm 0.24\) &
            \(\mathbf{87.17 \pm 0.85}\) \\

        ResNet-50 &
            \(80.67 \pm 0.62\) &
            \(\mathbf{82.50 \pm 0.41}\) \\

        \midrule
        \multicolumn{3}{c}{\(k\)-NN (\%)} \\
        \midrule
        Model & \multicolumn{1}{c}{Baseline} & \multicolumn{1}{c}{IDS} \\

        \midrule

        CNN-8\(\times\)8 & 
            \(64.67 \pm 0.24\) &
            \(\mathbf{71.17 \pm 0.24}\) \\

        CNN-16\(\times\)16 &
            \(64.17 \pm 0.24\) &
            \(\mathbf{71.50 \pm 1.78}\) \\

        CNN-32\(\times\)32 & 
            \(65.00 \pm 1.08\) &
            \(\mathbf{74.00 \pm 1.08}\) \\

        CNN-64\(\times\)64 &
            \(65.00 \pm 0.00\) &
            \(\mathbf{71.17 \pm 0.62}\) \\

        CNN-128\(\times\)128 &
            \(63.83 \pm 1.25\) &
            \(\mathbf{72.17 \pm 0.62}\) \\

        CNN-256\(\times\)256 &
            \(64.50 \pm 1.22\) &
            \(\mathbf{77.00 \pm 1.08}\) \\

        ResNet-18 &
            \(82.67 \pm 0.94\) &
            \(\mathbf{84.50 \pm 0.41}\) \\

        ResNet-50 &
            \(76.33 \pm 0.47\) &
            \(\mathbf{80.83 \pm 0.62}\) \\
        \bottomrule
    \end{tabular}
\end{table}

Table~\ref{tab:topline} reports linear-probe and \(k\)-NN accuracy on the
synthetic meteor dataset across the CNN and ResNet encoders. On the CNN
backbones, IDS matches or outperforms the flip-and-rotation baseline at every
input resolution under both probes, with the largest gains appearing at the
higher resolutions. The pattern reverses on the ResNet backbones, where the
in-domain FFT augmentation largely yields the best performance and IDS
underperforms the flip-and-rotation baseline, particularly on ResNet-18 under
the linear probe. We return to this in Section~\ref{sec:discussion}.

On CIFAR-10 (Table~\ref{tab:cifar10}), IDS improves over the flip-and-rotation
baseline on seven of eight encoder–probe combinations. We emphasize that this
comparison is deliberately restricted: to maintain parity with the meteor radar
augmentation protocol, both conditions use only flips and rotations rather than the full
SimCLR augmentation suite~\cite{cifar10}, which includes random resized crop,
color jitter, and grayscale conversion. Under that stronger protocol, canonical
SimCLR augmentations are known to substantially outperform flip-and-rotation
alone on CIFAR-10, and we make no claim that IDS competes with a well-tuned
data-space augmentation pipeline on natural imagery. The CIFAR-10 result is
therefore best read as evidence that IDS is not specific to the radar
domain---\ie, that weight-space perturbation produces useful view variation
under matched augmentation budgets---rather than as evidence that IDS is
preferable to data-space augmentation where the latter is physically
well-motivated.

Figure~\ref{fig:sweep-cnn16x16} sweeps the perturbation scale \(s_l\) on
CNN-16\(\times\)16 from $0.01$ to $0.20$. Performance is broadly insensitive
to the choice of \(s_l\) across this range: the linear probe varies between
$0.78$ and $0.82$ and the $k$-NN probe between $0.77$ and $0.80$, with n
sharp optimum. We did not observe the training instability predicted at
large \(s_l\) within the swept range, suggesting that for this architecture
the useful range of \(s_l\) extends well beyond the value used in
Tables~\ref{tab:topline} and~\ref{tab:cifar10}.

\section{Discussion}\label{sec:discussion}

The meteor dataset results support the central claim of the paper in a restricted
form. On the CNN encoders, replacing data-space augmentations with
weight-space perturbation produces representations of comparable or better
linear separability and local clustering, without requiring augmentations
that may corrupt the underlying physical signal. The CIFAR-10 results
indicate that this behavior is not specific to the radar domain.

The ResNet results on meteor radar are the clearest qualification. Applying IDS at
a single fully-connected layer after a much deeper convolutional stack
appears to under-perturb the representation; the bulk of the encoder's
capacity sits upstream of the perturbation site, and the resulting view
variation is presumably too small to drive the contrastive objective. This
is consistent with the design choice---perturbing one layer to keep the
budget comparable across architectures---failing to scale with depth. A
natural next step is to either widen the set of perturbed layers in deeper
backbones or to scale \(s_l\) with the receptive field downstream of the
perturbation site. This is left to future work.

The flatness of the \(s_l\) sweep is also worth noting. The theoretical
account in Section~\ref{sec:methodology} predicts degeneracy at \(s_l = 0\)
and instability at large \(s_l\), with a useful regime in between. We observe
the lower bound but not the upper, at least on CNN-16\(\times\)16 up to
\(s_l = 0.20\). Whether this reflects a property of the architecture, the
dataset, or the single-layer perturbation site cannot be easily discerned from the present experiments.

\section{Conclusion}\label{sec:conclusion}

We have presented implicit data synthesis as a contrastive augmentation
strategy that operates in weight space rather than data space, and
evaluated it against flip-and-rotation baselines on synthetic meteor radar
observations and on CIFAR-10. On the shallow CNN encoders, IDS matches or
outperforms the baseline on both datasets. On the deeper ResNet backbones
applied to meteor data, single-site perturbation is insufficient and the baseline
augmentations remain preferable. The technique is most useful in the
setting it was designed for: domains where data-space augmentations are
not physically well-motivated. Extending the analysis to multi-site
perturbation, to additional scientific imaging modalities, and to a
characterization of the upper bound on \(s_l\) are clear directions
for follow-up work.

\section*{Acknowledgments}

The authors graciously thank Sigrid Elschot and Nicolas Lee for their efforts
towards experiment planning and data analysis at Stanford University, Marco
Milla and Karim Kuyeng Ruiz for their assistance in data collection at
Jicamarca Radio Observatory, and Lindsey Marinello at the Georgia Institute of
Technology for her generous help with the manuscript. This work was supported
by NSF Grants 1920383, 2048349, and 2301645.



\bibliographystyle{abbrvnat}
\bibliography{references}

\appendix

\section{Technical appendices and supplementary material}\label{sec:technical-appendix}.

All experiments were run on a single NVIDIA DGX Spark, and took approximately seven days to complete.

\end{document}